\documentclass{article}

\PassOptionsToPackage{numbers, compress}{natbib}

\usepackage[preprint]{neurips_2022}
\usepackage{natbib}
\usepackage[utf8]{inputenc} 
\usepackage[T1]{fontenc}    
\usepackage{hyperref}       
\usepackage{url}            
\usepackage{booktabs}       
\usepackage{amsfonts}       
\usepackage{nicefrac}       
\usepackage{microtype}      
\usepackage{xcolor}         
\usepackage{bm}
\usepackage{graphicx}
\usepackage{amsmath}
\usepackage{algorithm}
\usepackage[noend]{algpseudocode}

\title{Fast Gaussian Process Posterior Mean Prediction via Local Cross Validation and Precomputation}

\author{%
  Alec M.~Dunton \\
  Center for Applied Scientific Computing\\
  Lawrence Livermore National Laboratory\\
  Livermore, CA 94550 \\
  \texttt{dunton1@llnl.gov} \\
  \And
  Benjamin W. Priest \\
  Center for Applied Scientific Computing\\
  Lawrence Livermore National Laboratory\\
  Livermore, CA 94550 \\
  \texttt{priest2@llnl.gov} \\
  \And
  Amanda Muyskens \\
  Applied Statistics Group \\
  Lawrence Livermore National Laboratory\\
  Livermore, CA 94550 \\
  \texttt{muyskens1@llnl.gov} \\
}

\begin{document}

\maketitle

\begin{abstract}
  Gaussian processes (GPs) are Bayesian non-parametric models useful in a myriad of applications.
  Despite their popularity, the cost of GP predictions (quadratic storage and cubic complexity with respect to the number of training points) remains a hurdle in applying GPs to large data. 
  We present a fast posterior mean prediction algorithm called FastMuyGPs to address this shortcoming.
  FastMuyGPs is based upon the MuyGPs hyperparameter estimation algorithm and utilizes a combination of leave-one-out cross-validation, batching, nearest neighbors sparsification, and precomputation to provide scalable, fast GP prediction. 
  We demonstrate several benchmarks wherein FastMuyGPs prediction attains superior accuracy and competitive or superior runtime to both deep neural networks and state-of-the-art scalable GP algorithms.
\end{abstract}

\section{Introduction and related work}
Gaussian processes (GPs) are popular non-parametric function approximation models found across spatial statistics~\cite{gelfand2010handbook,stein1999interpolation}, computer experiments~\cite{sacks1989design}, and machine learning applications~\cite{liu2020gaussian,williams2006gaussian}. 
GPs have fewer parameters and typically require far fewer training iterations than neural networks, and their posterior variance guards against overconfident prediction outside the training set.
Moreover, GPs are known to be sample efficient, giving them an advantage over highly parametrized approaches like deep neural networks.
However, GPs are not without drawbacks.
Predicting function values outside a training set with $n$ observations requires storing and inverting an $n \times n$ covariance matrix.
These operations require $\mathcal{O}(n^2)$ memory and $\mathcal{O}(n^3)$ computation.
This makes conventional GP prediction intractable for large-scale problems, e.g., datasets where $n \gg 10^5$.

The need to apply GPs to large data has prompted a proliferation of scalable GP algorithms.
The majority of these methods are approximation-based approaches that seek to produce a solution close to that of a standard GP~\cite{liu2020gaussian}. 
One can further sort approximation-based methods into local and global approaches.
Local approximations use a combination of GPs across the training data domain. 
The most basic implementation of a local approximation involves partitioning the training data and training separate GPs on each subdomain~\cite{fuentes2002spectral,gramacy2007tgp,sang2011covariance}.
These approaches can yield discontinuous results at the boundaries of the subdomain and generalize poorly due to their inability to detect global patterns in the response.
The mixture-of-experts approach combines the locally trained GPs by constructing Gaussian mixture models~\cite{jacobs1991adaptive}.
Product-of-expert approaches construct the likelihood as the product of all local processes instead of a weighted sum thereof~\cite{hinton2002training}.

Global approximations, which are the focal point of this paper, rely on sparsifying the kernel matrix. 
This can take the form of subsampling the data~\cite{hayashi2020random,lawrence2003fast,seeger2003bayesian,keerthi2006matching}, using kernels with compact support~\cite{gneiting2002compactly,melkumyan2009sparse,buhmann2001new,wendland2004scattered}, and sparse approximations inspired by low-rank methods such as the Nystr\"om approximation~\cite{liu2020gaussian}.
Another popular global approximation approach, covariance tapering, entails setting kernel matrix entries to zero if they are sufficiently small~\cite{furrer2006covariance,furrer2010spam}.

Inducing point methods construct a generative probabilistic model using a small set of latent variables to train the model.
These methods are among the most popular global approximations and include many successful variants.
Variational inference methods approximate the posterior~\cite{gibbs2000variational,lazaro2011variational,tran2015variational}.
Variational approaches are scalable; they have been successfully applied to problems with up to billions of data points~\cite{peng2017asynchronous,rivera2017forecasting}.
Structured kernel interpolation (SKI) methods~\cite{wilson2015kernel} exploit the kronecker structure of product kernels to reduce computational complexity for regressing $n$ training points from $\mathcal{O}(n^3)$ to as low as $\mathcal{O}(n)$~\cite{wilson2015kernel,wilson2016deep}. 
Spectral methods are another set of global approximation approaches that are well-suited for gridded data~\cite{guinness2019spectral,muyskens2018non}. 
Finally, some approaches assume sparsity in the precision matrix, i.e. the inverse of the kernel matrix~\cite{nychka2015multiresolution,lindgren2011explicit,datta2016hierarchical,vecchia1988estimation}.

Global approximations tend to model low frequency trends well, whereas local approximations tend to better model high frequency trends.
So-called hybrid approximation algorithms seek to balance the benefits and drawbacks of global and local approximations, e.g.,~\cite{lee2017hierarchically,snelson2007local}.
GPU-accelerated methods are another recent advance in scalable GPs~\cite{wang2019exact}. 
These approaches exploit kernel structure to parallelize tasks, leading to speedup in training and prediction time~\cite{dai2014gaussian,gal2014distributed}.
Incorporation of GPU hardware has led to the development of open source software libraries such as {\it GPflow}~\cite{matthews2017gpflow} and {\it GPyTorch}~\cite{gardner2018gpytorch}.
Dimensionality reduction-based approaches such as manifold GP address cases where the feature space is large~\cite{lawrence2005probabilistic,snelson2012variable}.
Other recent contributions to scalable GP regression include deep GPs~\cite{wilson2016deep,damianou2013deep}, online GPs~\cite{csato2002sparse}, and recurrent GPs~\cite{al2017learning}.

In this work we focus on the fast, online GP posterior mean prediction problem. 
This problem is relevant to, e.g., Reinforcement Learning (RL) and inverse problems.
In RL and inverse problems the given model is assumed to be trained, and having access to rapid successive model evaluations is critical.
We build upon the MuyGPs hyperparameter optimization algorithm to enable fast prediction.
MuyGPs has achieved state-of-art accuracy, memory usage, and training time results on common GP benchmarks~\cite{heaton2019case,muyskens2021muygps}.
However, posterior mean prediction remains a computational bottleneck in the existing MuyGPs workflow.
This work directly addresses this shortcoming and describes an end-to-end scalable GP regression algorithm.

The method proposed in this work, FastMuyGPs, augments MuyGPs by approximately pre-computing parts of the GP prediction equation as a part of the training procedure, reducing the prediction time of the algorithm.
The resulting approach reduces the floating-point operation (FLOP) complexity of predicting a test point using standard GPs from $\mathcal{O}(n^3)$ to $\mathcal{O}(\log(n) + k)$, where $k$ is the number of nearest neighbors used to form a prediction.
FastMuyGPs incurs a memory cost of $\mathcal{O}(nk)$ as opposed to the $\mathcal{O}(n^2)$ memory required to store the covariance matrix for $n$ training points. 
In our experiments, FastMuyGPs achieves faster prediction times and superior accuracy compared to a selection of scalable GP approaches implemented using the GPyTorch library~\cite{gardner2018gpytorch}.
Moreover, the approach competes directly with neural networks and GPU-optimized GP prediction algorithms like SKI in prediction time, further demonstrating its efficacy. 

The rest of this paper is organized as follows. 
Section~\ref{sec:background} provides relevant background on GPs.
Section~\ref{sec:fastmuygps} describes FastMuyGPs. 
Section~\ref{sec:experiments} demonstrates our method on three test problems, benchmarked against several state-of-the art scalable GP regression approaches implemented in GPyTorch, as well as deep neural networks (DNNs) implemented in PyTorch.
Section~\ref{sec:future} draws conclusions on FastMuyGPs and suggest future avenues of research.
Section~\ref{sec:society} provides a discussion on the broader societal impacts of our work.
The Supplementary Material provides additional details on the computing resources, training procedures, and datasets used in our experiments. 

\section{Background}\label{sec:background}
In the standard setting, Gaussian process regression (GPR) seeks to approximate a function $f : \mathbb{R}^{d} \rightarrow \mathbb{R}$ given $n$ training points in $\bm{R}^d$ constituting the rows of a matrix $\bm{X} \in \mathbb{R}^{n \times d}$, 
and corresponding observation vector $Y(\bm{X}) \in \mathbb{R}^{\ell}$.
We assume that $f$ is drawn from the  distribution $\mathcal{GP}(m(\bm{x}),k_\theta(\bm{x},\bm{x'}))$.
In this expression, $m(\bm{x})$ denotes the mean of the GP evaluated at the location $\bm{x}$. 
We typically set the mean function $m(\bm{x})$ to zero after the data has been de-trended by, e.g., polynomial regression.
$k_\theta(\bm{x},\bm{x'})$ is the {\it kernel} function, which generates the covariance between $\bm{x}$ and $\bm{x'}$ and is controlled by hyperparameters $\theta$.
The exponential, radial basis function (RBF), and Mat\'ern kernels are common in applications~\cite{williams2006gaussian}.

The training data responses $Y(\bm{X})$ can be thought of as a finite sample of a stochastic process $Y$. 
We call $Y$ a Gaussian process if every finite sample of $Y$ is multivariate Gaussian distributed, i.e.
\begin{equation}
Y(\bm{X}) \sim \mathcal{N}(\bm{m}(\bm{X}),\bm{K}_{\theta}(\bm{X},\bm{X})). 
\end{equation}
Here $\mathcal{N}$ denotes the multivariate Gaussian distribution, $\bm{m}(\bm{X})$ is the mean vector of size $n$, 
and $\bm{K}_{\theta}(\bm{X},\bm{X})$ is the covariance matrix generated by the kernel $k(\bm{x},\bm{x'})$ with hyperparameters $\theta$.
We use the conditional distribution of the response at new points $\bm{Z}$ given the observed training data $\bm{X}$ to predict the response $Y(\bm{Z})$ with $\hat{Y}(\bm{Z} \vert \bm{X})$:
\begin{equation}
\hat{Y}(\bm{Z} \vert \bm{X}) = \bm{K}_{\theta}(\bm{Z},\bm{X})\bm{K}_{\theta}{(\bm{X},\bm{X})}^{-1}Y(\bm{X}).
\end{equation}
Here $\bm{K}_{\theta}(\bm{Z},\bm{X})$ is the cross covariance of the test points $\bm{Z}$ and training data $\bm{X}$.
We call the matrix $\bm{K}_{\theta}(\bm{Z},\bm{X})\bm{K}_{\theta}{(\bm{X},\bm{X})}^{-1}$ the {\it Kriging weights}, which can be thought of as imposing a weighted average on the training responses $Y(\bm{X})$. 
The conditional mean $\hat{Y}(\bm{Z} \vert \bm{X})$ is known to be the best linear unbiased predictor of ${Y}(\bm{Z})$~\cite{santner2003design}. 

The log-likelihood of the training data $Y(\bm{X})$ given $\theta$ is defined as 
\begin{equation}
  \text{log}(L(\theta,Y(\bm{X}))) = -\frac{p}{2}\text{log}(2\pi) - \frac{1}{2}\text{log}(\bm{K}_{\theta}(\vert \bm{X},\bm{X})\vert) - \frac{1}{2} {Y(\bm{X})}^T \bm{K}_{\theta}{(\bm{X},\bm{X})}^{-1}Y(\bm{X}) .
  \label{eqn:likelihood_func}
\end{equation}
Conventional GP training consists of maximizing this quantity with respect to $\theta$.
Note that computing the determinant and linear solve in the likelihood function requires $\mathcal{O}(n^3)$ FLOPs and $\mathcal{O}(n^2)$ storage.
This is prohibitively expensive in large-scale applications, necessitating either approximations to the log-likelihood or the use of a different objective function to train $\theta$.

Applications often assume that the covariance matrix $\bm{K}_{\theta}{(\bm{X},\bm{X})}$ is induced by a stationary and isotropic kernel.
That is, the kernel function satisfies $k_\theta(\bm{x},\bm{x'}) = \phi_\theta(\Vert \bm{x} - \bm{x'} \Vert)$, where $\phi :  \mathbb{R}_{\geq 0} \rightarrow \mathbb{R}_{\geq 0}$ is a function parametrized by $\theta$.
The Mat\'ern kernel function is a primary example of a stationary and isotropic kernel.
Let $d = \Vert \bm{x} - \bm{x'} \Vert$, the distance between a pair of points $\bm{x}$ and $\bm{x'}$. 
The Mat\'ern kernel is defined as
\begin{equation}
  \phi_{\sigma,\rho,\nu,\tau}(d) = \sigma^2 \left[ \frac{2^{1-\nu}}{\Gamma(\nu)} \left( \sqrt{2\nu}\frac{d}{\rho} \right)^{\nu} K_{\nu} \left( \sqrt{2\nu}\frac{d}{\rho} \right) + \tau^2\mathbb{I} \left( d = 0 \right) \right] ,
\end{equation}
where the hyperparameters are $\lbrack \sigma,\rho,\nu,\tau \rbrack$, $\Gamma(\nu)$ is the gamma function evaluated at $\nu$, and $K_{\nu}$ is the modified Bessel function of the second kind.  
At $\nu = 1/2$, the Mat\'ern kernel is equal to the exponential kernel, and as $\nu \rightarrow \infty$, the Mat\'ern kernel converges pointwise to the RBF kernel. 
For small values of $\nu$, the Mat\'ern kernel generates non-smooth predictions, whereas in the $\nu \rightarrow \infty$ limit, the Mat\'ern kernel produces smooth approximations. 
Throughout the experiments in Section~\ref{sec:experiments}, we employ the Mat\'ern kernel with different values of $\nu$.

\subsection{MuyGPs}
\begin{algorithm}[h]
  \caption{MuyGPs~\cite{muyskens2021muygps}}\label{alg:muygps}
    \begin{algorithmic}[1]
    \Procedure{MUYGPS$_{train}$}{$k$, $b$, $\bm{X}$, $Y(\bm{X})$, $\theta$}
    \State $b \leftarrow$ batch size
    \State $k \leftarrow$ number of nearest neighbors
    \State $\bm{X}, Y(\bm{X}) \leftarrow$ train features and responses 
    \State $\theta \leftarrow$ hyperparameters initial guess
    \State $B \leftarrow$ uniform sample of size b from $\lbrace 1, \dots , n \rbrace$
    \State $\bm{X}_{N_i} \leftarrow$ query $k$ nearest neighbors for all $i \in B$
    \State $Q(\theta) = \frac{1}{b} \sum_{i \in B} {\left(Y(\bm{x}_i) - \hat{Y}_{\theta}(\bm{x}_i \vert \bm{X}_{N_i})\right)}^2 $
    \State $\hat{\theta} \leftarrow \min_{\theta} Q(\theta)$ minimize loss function using, e.g., L-BFGS-B
    \State \textbf{return} $\hat{\theta}$ optimized hyperparameters for use in prediction
    \EndProcedure
    \end{algorithmic}
  \end{algorithm}
MuyGPs is a global approximation algorithm that accelerates hyperparameter optimization by limiting the kernel matrix to the nearest neighbor structure of the training data (thereby inducing sparsity on the Kriging weights), batching, and replacing expensive log-likelihood evaluations with leave-one-out cross validation (LOOCV). 
MuyGPs can scale prediction problems with millions of data points on a standard laptop computer~\cite{muyskens2021muygps}. 

LOOCV withholds the $i$th training location $\bm{x}_i$ and predicts its response $Y(\bm{x}_i)$ using the other $n-1$ points.
However, computing this prediction requires $\mathcal{O}(n^3)$ FLOPs, and therefore would not offer any complexity reduction relative to likelihood function evaluation.
MuyGPs reduces this complexity by conditioning $\mathbf{x}_i$ only on its $k$ nearest neighbors, denoted $\bm{X}_{N_i}$, yielding the prediction
\begin{align}
  \hat{Y}_{\theta}(\bm{x}_i \vert \bm{X}_{N_i}) &= \bm{K}_{\theta}(\bm{x}_i,\bm{X}_{N_i}) {\bm{K}_{\theta}(\bm{X}_{N_i},\bm{X}_{N_i})}^{-1}Y(\bm{X}_{N_i}).
\end{align}
The MuyGPs training procedure minimizes a loss function $Q(\theta)$ over a randomly sampled batch of training points $B$ with $b = \vert B \vert$.
In this work we define $Q(\theta)$ in terms of mean squared error.
Training $\theta$ amounts to minimizing $Q(\theta)$ with respect to $\theta$:
\begin{equation} \label{eqn:q}
  Q(\theta) = \frac{1}{b} \sum_{i \in B} {\left(Y(\bm{x}_i) - \hat{Y}_{\theta}(\bm{x}_i \vert \bm{X}_{N_i})\right)}^2; \hspace{24pt} \hat{\theta} = \min_{\theta} Q(\theta).
\end{equation}
Evaluating Equation~(\ref{eqn:q}) requires $\mathcal{O}(bk^3)$ FLOPS.
This is much cheaper than the $\mathcal{O}(n^3)$ cost of the log-likelihood in Equation~(\ref{eqn:likelihood_func}) if $b \ll n$ and $k \ll n$.
MuyGPs predicts $Y(\bm{z})$ for a novel point $\bm{z}$ with nearest neighbors in $\mathbf{X}$ denoted $\bm{X}_{N^*}$ via
\begin{equation}
  \hat{Y}_{\hat{\theta}}(\bm{z} \vert \bm{X}) = \bm{K}_{\hat{\theta}}(\bm{z},\bm{X}_{N^*})\bm{K}_{\hat{\theta}}{(\bm{X}_{N^*},\bm{X}_{N^*})}^{-1}Y(\bm{X}_{N^*}) \label{eqn:muygps_mean}.
\end{equation}

\section{FastMuyGPs}\label{sec:fastmuygps}
\begin{algorithm}[h]
  \caption{FastMuyGPs}\label{alg:fast_GP_pred}
    \begin{algorithmic}[1]
    \Procedure{FASTMUYGPS}{$k$, $b$, $\bm{X}$, $Y(\bm{X})$, $\theta$}
    \State $b \leftarrow$ batch size
    \State $k \leftarrow$ number of nearest neighbors
    \State $\bm{X}, Y(\bm{X}) \leftarrow$ train features and responses
    \State $\theta \leftarrow$ hyperparameters initial guess
    \State \textbf{Offline Training}
    \State $\hat{\theta} \leftarrow \text{MuyGPs}(k,b, \bm{X}, Y(\bm{X}), \theta)$
    \State \textbf{for} i = 1: $n_{train}$
    \State \qquad $\bm{X}_{N_i} \leftarrow$ query $k - 1$ nearest neighbors of $\bm{x}_i$
    \State \qquad $\bm{X}_{S_i} = \lbrack \bm{x}_i ; \bm{X}_{N_i} \rbrack$
    \State \qquad $\bm{C}_i = {\bm{K}_{\hat{\theta}}(\bm{X}_{S_i},\bm{X}_{S_i})}^{-1}Y(\bm{X}_{S_i})$
    \State \textbf{end}
    \State \textbf{Online Prediction}
    \State $\bm{Z} \leftarrow$ test features 
    \State \textbf{for} t = 1: $n_{text}$
    \State \qquad $\bm{X}_{N_t^*} \leftarrow$ query nearest training point and its $k-1$ nearest neighbors $\forall \bm{z}_t \in \bm{Z}$\label{algstep:test_nearest_neighbor}
    \State \qquad $\tilde{Y}_{\hat{\theta}}(\bm{z}_t \vert \bm{X}) \leftarrow \bm{K}_{\hat{\theta}}(\bm{z}_t,\bm{X}_{N_t^*})\bm{C}_{N_t^*}$
    \State \textbf{end}
    \State \textbf{return} $\tilde{Y}_{\hat{\theta}}(\bm{Z} \vert \bm{X})$ 
    \EndProcedure
    \end{algorithmic}
  \end{algorithm}
FastMuyGPs accelerates prediction by precomputing the part of Equation~(\ref{eqn:muygps_mean}) that depends only on the training data during the training procedure.
This reduces the cost of predicting a single test point from $\mathcal{O}(n^2)$ to $\mathcal{O}(\log(n) + k)$.
Following MuyGPs hyperparameter training, we form a $n \times k$ matrix, which we denote $\bm{C}$.
Let ${S_i} =  i \cup {N_i} $, i.e., the union of $i$ and the indices of the $k-1$ nearest neighbors of training point $i$. 
Then, the $i^{th}$ row of $\bm{C}$ is defined as
\begin{align}
  \bm{C}_i &= {\bm{K}_{\hat{\theta}}(\bm{X}_{S_i},\bm{X}_{S_i})}^{-1}Y(\bm{X}_{S_i}).
\end{align}
$\bm{C}$ depends exclusively on the training data, so we can precompute it as a part of the training procedure.
For a test point $\bm{z}$ with the index set $N^*$ containing $\bm{z}$'s nearest neighbor and that neighbor's $k-1$ nearest neighbors and corresponding training features $\bm{X}_{N^*}$, we employ the following formula to estimate $Y(\bm{z})$:
\begin{align}
  \tilde{Y}_{\hat{\theta}}(\bm{z} \vert \bm{X}) &= \bm{K}_{\hat{\theta}}(\bm{z},\bm{X}_{N^*})\bm{C}_{N^*}.
\end{align}
Here $\bm{C}_{N^*}$ denotes the matrix consisting of the rows of $\bm{C}$ corresponding to the indices contained in $N^*$, i.e., the precomputed coefficients whose dot product with the test-training cross-covariances produces the prediction. 

Predicting the response of a single test point in FastMuyGPs consists of a nearest neighbor's coefficient lookup, forming the cross-covariance vector, and applying a vector-vector dot product. 
Approximate nearest neighbor lookups require only $\log(n)$ time using fast algorithms, e.g., hierarchical navigable small worlds (HNSW)~\cite{malkov2018efficient}. 
For $\mathcal{O}(m)$ test points and $\mathcal{O}(n)$ training points using $k$ nearest neighbors to form the approximation, we obtain a total prediction complexity $\mathcal{O}(m(\log(n) + k))$.
This is an improvement over the $\mathcal{O}(m(\log(n) + k^3))$ prediction complexity of MuyGPs, and a vast improvement over the $\mathcal{O}(n^3 + mn^2)$ complexity of standard GP regression.

SKI-based methods, the fastest prediction methods in GP regression, incur a superior $\mathcal{O}(m)$ prediction cost with precomputing.
If we interpret $k$ to be a fixed parameter like the interpolation weights in SKI are, then FastMuyGPs has complexity worse only by a multiplicative factor of $\log(n)$.
However, FastMuyGPs achieves far better accuracy than SKI in the three benchmark problems presented in Section~\ref{sec:experiments}.
This may be because FastMuyGPs sparsifies the kernel matrix via local Kriging, as opposed to the inducing point interpolation approach used in SKI.
Moreover, FastMuyGPs has a $\mathcal{O}(nk)$ memory overhead; a similar precomputation method using standard GP regression would feature $\mathcal{O}(n^2)$ overhead. 
Other scalable GP approaches have more favorable memory requirements than FastMuyGPs; we leave reducing the memory footprint of FastMuyGPs to a future work.

\section{Experiments}\label{sec:experiments}
\begin{figure}[bh!p]
  \centering
  \includegraphics[width=\linewidth]{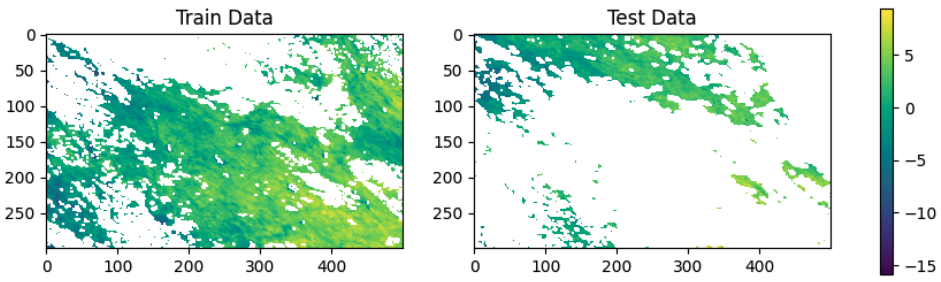}
  \caption{Normalized surface temperatures (originally in Celsius) measured on August 4, 2016 between longitudes -95.91153 and -91.28381 and latitudes 34.29519 to 37.06811 collected by a Terra instrument from the MODIS satellite.}\label{fig:heaton_snaps}
  \end{figure}

We benchmark the performance of FastMuyGPs  (Algorithm~\ref{alg:fast_GP_pred}) against fast GP algorithms implemented as a part of the GPyTorch library, as well as deep neural networks implemented using PyTorch.
Our first test case is a spatial statistics regression problem\footnote{https://github.com/finnlindgren/heatoncomparison} previously used to compare scalable GP methods~\cite{heaton2019case}.
The second test case is a classic inverse problem, where the parameters of a function modeling the outflow of a borehole are estimated with a Markov Chain Monte Carlo (MCMC) calibration assisted by GP emulation.\footnote{https://www.sfu.ca/ssurjano/borehole.html}
The third and final test case is a binary star-galaxy image classification problem inspired by previous applications of MuyGPs to astronomy~\cite{buchanan2022gaussian,muyskens2022star}.\footnote{https://github.com/Kerianne28/}
Throughout this section, execution timings are measured on an 8GB NVIDIA Tesla P100 GPU.

\subsection{FastMuyGPs as a predictor: surface temperature prediction}

The dataset we study in this test case comprises land surface temperatures measured by a Terra instrument from the MODIS satellite on August 4, 2016.
We measure the quantity-of-interest (QoI) on a $500 \times 300$ grid between longitudes -95.91153 and -91.28381 and latitudes 34.29519 to 37.06811.
We impose the dataset's train-test split via an artificial cloud cover pattern drawn from a different measurement interval.
Figure~\ref{fig:heaton_snaps} depicts the 105,569 training observations and 42740 testing observations left over after removing unmeasured points.

We benchmark FastMuyGPs against three fast, scalable GP implementations from the GPyTorch software library: SKI~\cite{wilson2015kernel}, sparse GPR (SGPR)~\cite{titsias2009variational}, and stochastic variational GPR (SVGP)~\cite{hensman2015scalable}. 
We also include comparisons to a fully-connected regression neural network (FCNN) to demonstrate the speed of our predictor.
For the four benchmark approaches, we withhold 10557 training samples for validation to select the best model during training. 

\begin{figure}[h!]
  \includegraphics[width=0.49\linewidth]{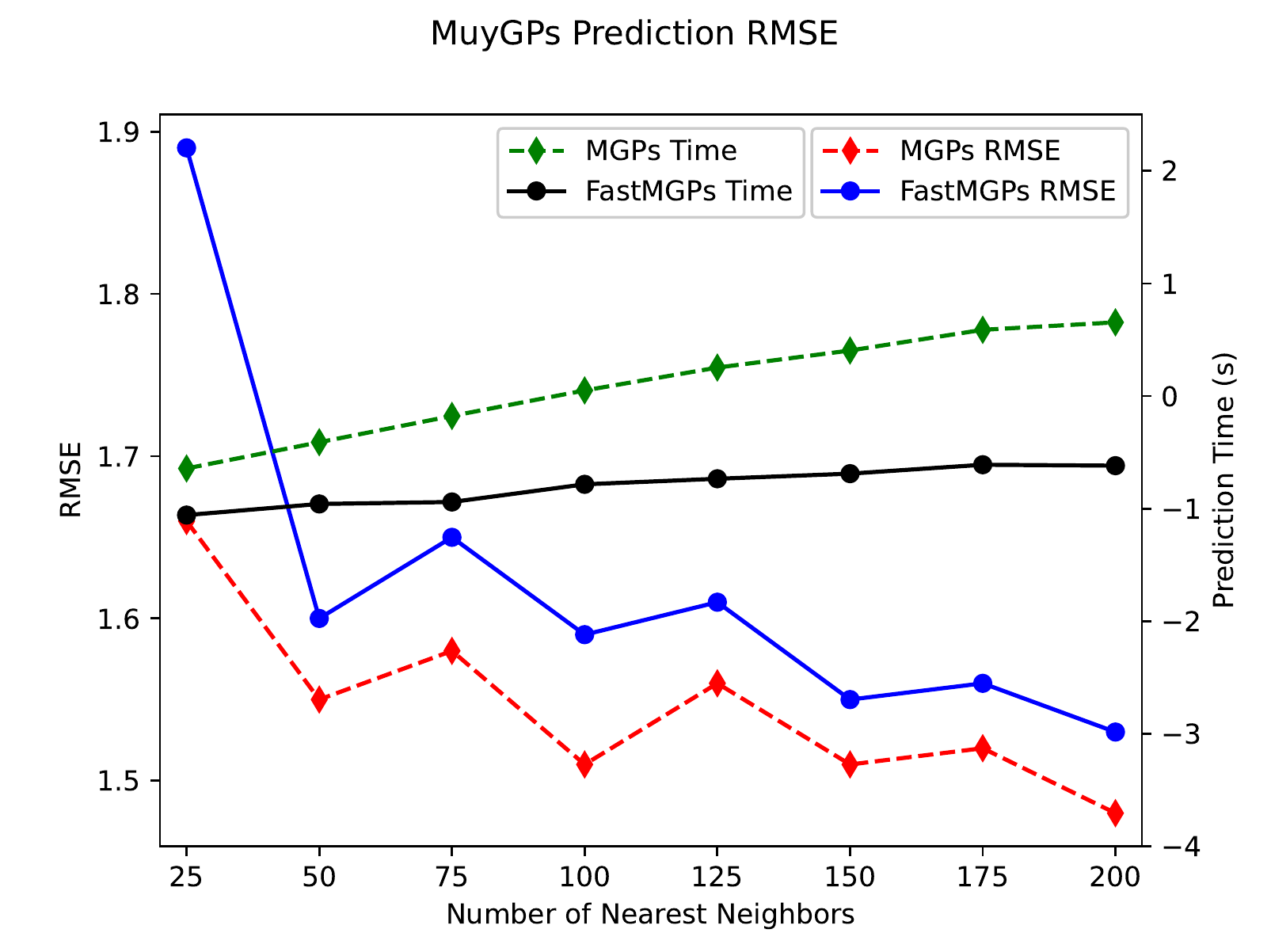}
  \includegraphics[width=0.49\linewidth]{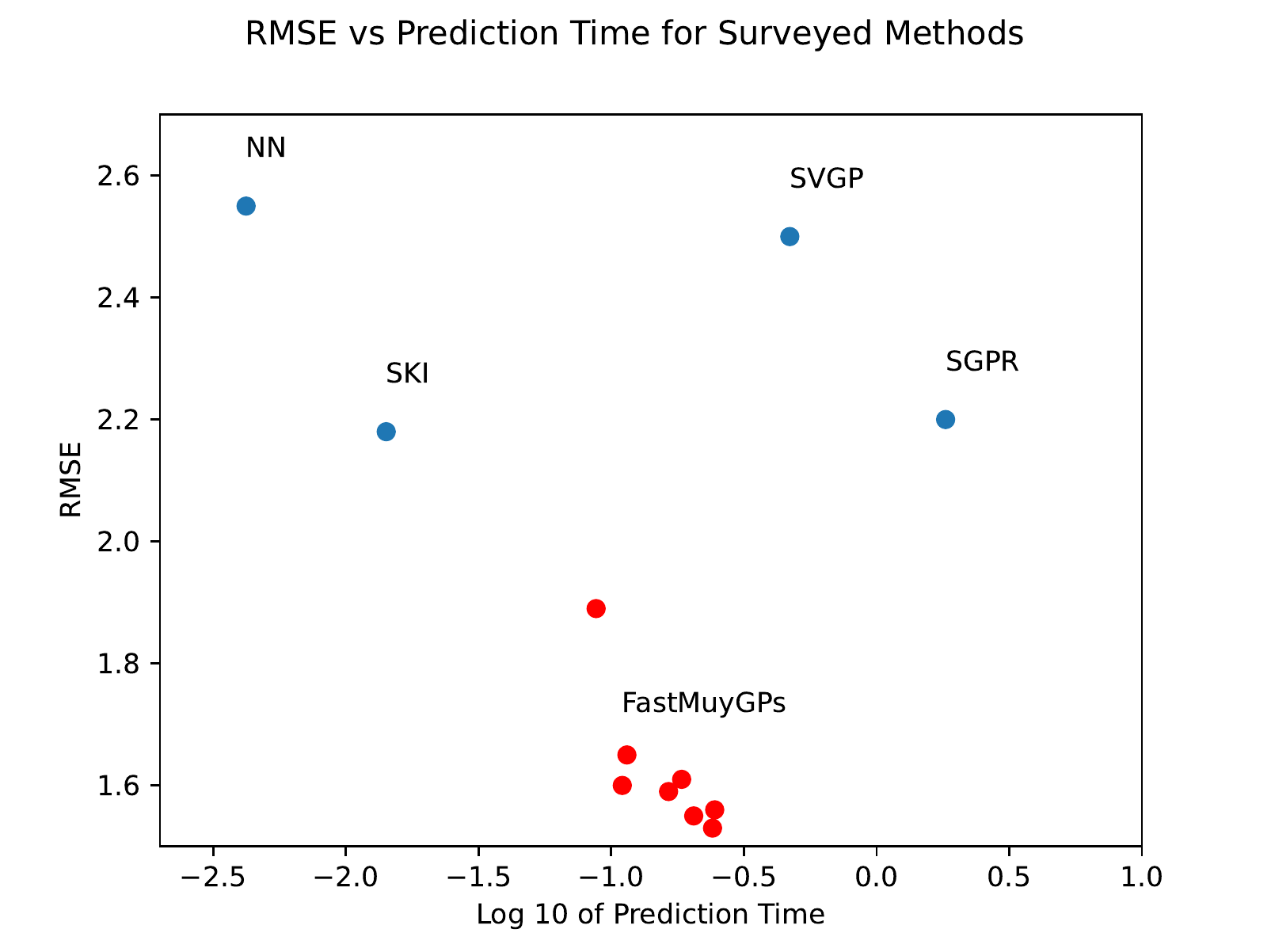}
  \caption{Left: Comparisons of root-means squared error (RMSE) and prediction time for MuyGPs and FastMuyGPs on the temperature dataset for different numbers of nearest neighbors. 
           Right: (RMSEs, y-axis) and log base 10 of the prediction times (x-axis) of SKI~\cite{wilson2015kernel}, SGPR~\cite{titsias2009variational}, SVGP~\cite{hensman2015scalable}, and FastMuyGPs.}\label{fig:times_rmse_heaton}
  \end{figure}

Examining the left panel of Figure~\ref{fig:times_rmse_heaton}, we see that FastMuyGPs is slightly less accurate than MuyGPs, while offering improved scaling with respect to the number of nearest neighbors.
The approximation error increases because we approximate the test point using the nearest neighbors of the training point nearest to the test point, as opposed to the nearest neighbors of the test point itself.
The right panel of Figure~\ref{fig:times_rmse_heaton} benchmarks the performance of FastMuyGPs against four fast predictive approaches: FCNN, SKI, SVGP, and SGPR.
In the right panel of Figure~\ref{fig:times_rmse_heaton} we observe that FastMuyGPs achieves near the average prediction time among the methods, competing directly with SKI, which is known to be among the fastest GPR prediction schemes.
Moreover, FastMuyGPs is more accurate than the neural network, and both more accurate and faster than SPGR and SVGP. 
MuyGPs achieved state-of-the-art RMSE on this benchmark problem in~\cite{muyskens2021muygps}.
However, compared to the most accurate state-of-the-art methods in~\cite{heaton2019case}, the most accurate of which had an RMSE of 1.53, FastMuyGPs performs favorably.
Therefore, in this example, FastMuyGPs gives us the best of both worlds: fast prediction and near-state-of-the-art accuracy.

\subsection{FastMuyGPs as an emulator for MCMC: the borehole function inverse problem}

We study the borehole function - a popular benchmark for uncertainty quantification using emulation - in the second test case~\cite{harper1983sensitivity,worley1987deterministic,morris1993bayesian,simulationlib}. 
This test case represents the common inverse problem workflow where we train an expensive computer model on a set number of model evaluations. This number is many less than the evaluations needed in the MCMC calibration that uses a likelihood to compare predictions from the emulator to data.
We first show that FastMuyGPs offers a highly accurate approximation of the function. 
Then, we use FastMuyGPs as an emulator to estimate the width of the borehole based on the response (flow rate).
It is critical that we can generate accurate approximations of the ground truth rapidly to generate enough samples to estimate the input in this context.
We demonstrate that FastMuyGPs satisfies both of these requirements. 

We define the borehole function
\begin{equation}
  \label{eqn:borehole}
  f(r_w,r,T_u,H_u,T_l,H_l,L,K_w) = \frac{2\pi T_u (H_u - H_l)}{\ln(r/r_w)\left(1 + \frac{2LT_u}{\ln(r/r_w)r_w^2K_w} + \frac{T_u}{T_l}\right)} .
\end{equation}
$f(\cdot)$ is the rate of water flow given in $m^3/yr$ through a borehole. 
The 8 parameters are: $r_w \in \lbrack 0.05, 0.15 \rbrack$, the radius of the borehole ($m$); $r \in \lbrack 100, 50000 \rbrack$, the radius of influence ($m$); 
$T_u  \in \lbrack 63070, 115600 \rbrack$, the transmissivity of the upper aquifer ($m^2/yr$); $H_u \in \lbrack 990, 1110 \rbrack$, the potentiometric head of the upper aquifer ($m$); $T_l \in \lbrack 63.1, 116 \rbrack$, the transmissivity of the lower aquifer ($m^2/yr$);
$H_l \in \lbrack 700, 820 \rbrack$, the potentiometric head of the lower aquifer ($m$); $L \in \lbrack 1120, 1680 \rbrack$, the length of the borehole ($m$); and $K_w \in \lbrack 9855, 12045 \rbrack$, the hydraulic conductivity of the borehole ($m/yr$)~\cite{morris1993bayesian}.

We use the fast predictive capability of FastMuyGPs to recover the radius of the borehole $r_w$ from the flow $f(\cdot)$. 
We fix the parameters $H_u$, $H_l$, $L$, and $K_w$, while allowing the parameters $r_w$, $r$, $T_u$, and $T_l$ to vary. 
These variables must be fixed to ensure that the inverse problem is identifiable.
We generate the training and testing data using a Latin Hypercube (LHC) sampling strategy to produce data sampled in the four non-fixed variables.
We then pass the sampled values through an affine transformation to ensure that they are distributed between the bounds enumerated above.
We then divide the eight transformed parameters element-wise by the vector $\lbrack 0.0625, 0.25, 1, 0.25, 0.5, 0.25, 0.125, 0.5 \rbrack$ following the approach in~\cite{cole2021locally} to introduce anisotropy into the model.
Finally, we subtract the mean of the response variable from the training and test responses to enforce a zero mean on the process.

\begin{table}[h!]
  \centering
\begin{tabular}{ |p{2.3cm}|p{1.6cm}|p{1.7cm}|p{1.2cm}| p{1.2cm}| p{1.2cm}| p{1.2cm}|  }
  \hline
      & MuyGPs & FastMuyGPs & FCNN & SKIP & SVGP & SGPR \\
  \hline
  RMSE &  1.19e-2 & \textcolor{red}{$\bm{1.12e}$$\bm{-2}$}  &  4.65e-1  &  3.87e-1  & 4.37e-1    & 5.15e-1 \\
  \hline
  Predict Time (s)  &  1.32  &  1.09e-1 &\textcolor{red}{$\bm{4e}$$\bm{-3}$} & 4.41 & 6.07e-2 & 1.43e-1 \\
  \hline
 \end{tabular}
 \newline
 \caption{RMSE and prediction time for the selected approaches. Highlighting represents the best performance in each statistic.}
\label{table:borehole}
\end{table}

We train a MuyGPs model on 100000 training examples with 20000 test examples withheld.
We use an RBF Kernel with a learned length scale of 43.27 and 150 nearest neighbors to form the approximation.
We benchmark FastMuyGPs against scalable kernel interpolation for product kernels (SKIP), SVGP, SGPR, and FCNN  (we use SKIP in place of SKI because the problem is higher dimensional).
We withhold 10000 training samples for validation to select the best model during training for the four benchmark approaches.

In Table~\ref{table:borehole}, we see that the FastMuyGPs algorithm achieves accuracy superior to that of MuyGPs, as well as all other approaches.
MuyGPs achieves an RMSE of 1.19e-2, whereas FastMuyGPs achieves an RMSE of 1.12e-2; both of these results are on par with the state-of-the-art RMSE result on the problem presented in~\cite{cole2021locally}.
Moreover, FastMuyGPs achieves a speedup greater than 10X compared to MuyGPs.
FastMuyGPs also compares favorably to the fully-connected neural network and three GPyTorch methods.

\begin{figure}[h!]
  \centering
  \includegraphics[width=0.9\linewidth]{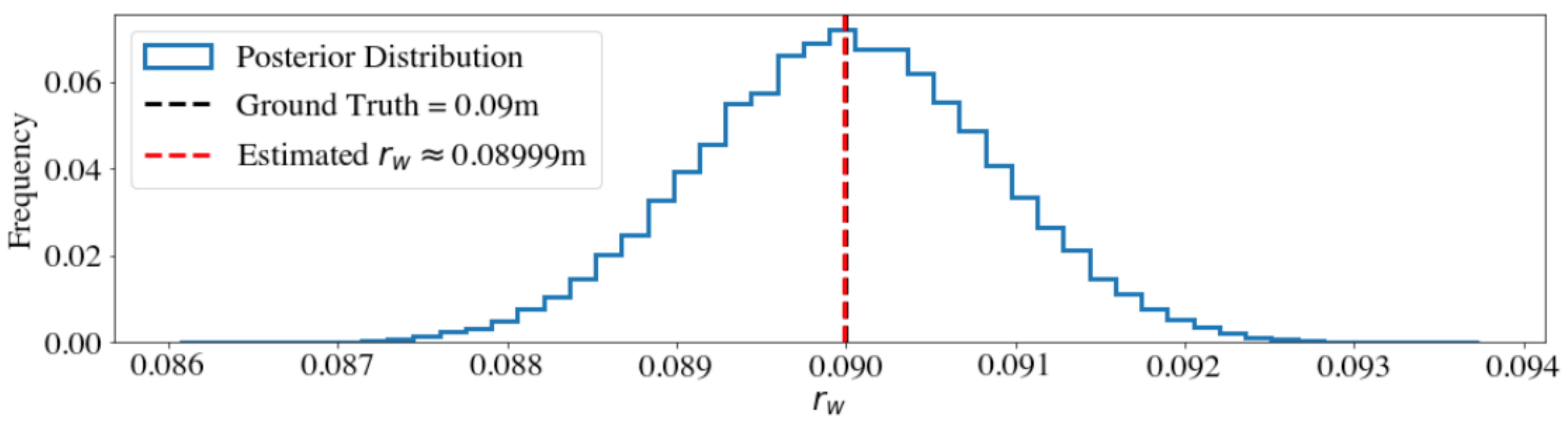}
  \caption{Posterior distribution of borehole radius ($r_w$) estimated using a FastMuyGPs GP emulator of the borehole function given in Equation~(\ref{eqn:borehole}). We display the ground truth as a dashed black line, and the posterior mean as a red dashed line.}
  \label{fig:borehole_mcmc}
  \end{figure}

We now demonstrate the utility of a fast predictive GP approach in the context of inverse problems.
We use an MCMC Algorithm to learn the parameter $r_w$ based on realizations of a GP model, with the parameters $H_u$, $H_l$, $L$, and $K_w$ fixed. 
The MCMC algorithm utilizes an acceptance probability that depends upon the prediction error (difference between the ground truth target value and response predicted by the GP).
We set the ground truth parameter $r_w$ to $0.09$, the standard deviation of the target distribution to 1, and run $10^7$ MCMC samples. 
In Figure~\ref{fig:borehole_mcmc}, we see that the MCMC accurately recovers the radius of the borehole given the observed flow rate, returning a posterior mean of 0.3998 as opposed to 0.4 - approximately 0.05 percent relative error.
The emulator learned by the FastMuyGPs predictor provides a scalable method for MCMC-based inverse problems.

\subsection{FastMuyGPs as a classifier: star-galaxy identification of ZTF images}

\begin{figure}[h!]
  \centering
  \includegraphics[width=0.8\linewidth]{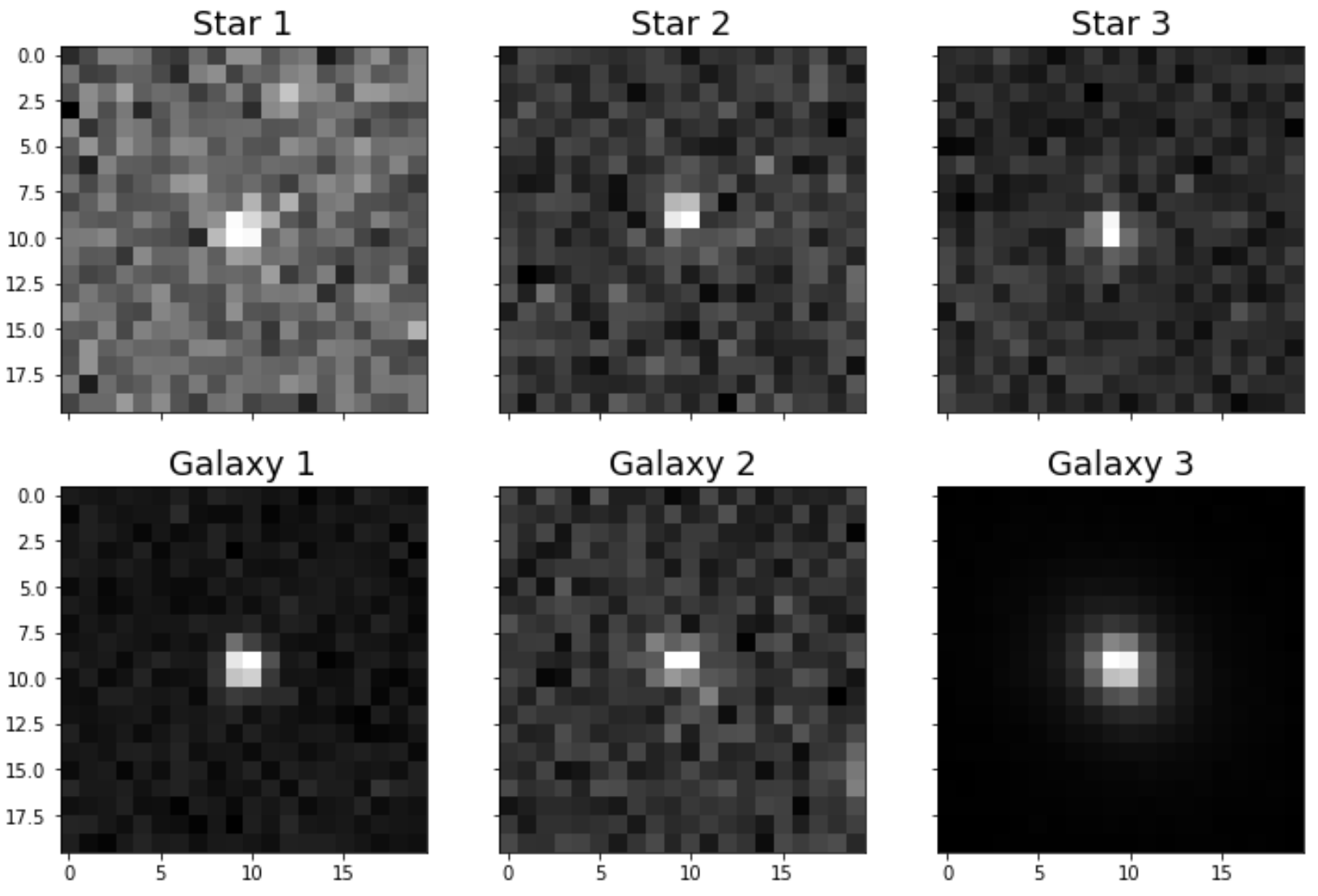}
  \caption{Sample of three images of stars and galaxies generated using code from~\cite{pruett2021ztf_image_classification}.} \label{fig:star_galaxy}
  \end{figure}

We examine a binary star-galaxy classification problem (see Figure~\ref{fig:star_galaxy}) as our third test case. 
Our dataset comprises roughly 50,000 $20 \times 20$ black-and-white images: approximately 25,000 of stars and 25,000 of galaxies extracted using the ZTF (Zwicky Transient Facility) image classification scripts~\cite{pruett2021ztf_image_classification,masci2018zwicky}.
We split the training and testing datasets 50/50, and train various classifiers to predict whether or not a given image is a star or a galaxy.
We benchmark our approach against four methods in this case: MuyGPs, a deep convolutional neural network (CNN), SVGP, and SGPR.
We withhold 2499 training samples for validation to select the best model during training for the three benchmark approaches.

  \begin{table}[h]
    \centering
  \begin{tabular}{ |p{2.3cm}|p{1.6cm}|p{1.7cm}|p{1.3cm}|p{1.6cm}| p{1.3cm}|  }
    \hline
        & MuyGPs & FastMuyGPs & CNN & SVGP & SGPR \\
    \hline
    Accuracy &  \textcolor{red}{$\bm{99.03}$} & 96.33 &  97.18   & 86.42   & 81.08 \\
    \hline
     Predict Time (s)   &  9.86e-1  & 1.41e-1 & \textcolor{red}{$\bm{3.23e}$$\bm{-2}$} &  2.16e-1 & 6.9e-2 \\
    \hline
   \end{tabular}
   \newline
   \caption{Accuracy and prediction time on the original, $400$-dimensional star-galaxy data. Highlighting represents the best performance in each statistic.}
  \label{table:star_galaxy_original}
  \end{table}

Examining Table~\ref{table:star_galaxy_original}, we see that the fastest of the methods is the CNN, followed by SGPR and FastMuyGPs.
FastMuyGPs achieves accuracy better than that of the two GPyTorch inducing point methods, but slightly lower than that of the CNN and MuyGPs. 
The speedup of FastMuyGPs over MuyGPs in this case is approximately 7X, which is less than that observed in previous test cases.
The relatively small gain in speed in this case is due to the computational cost of identifying the training point closest to a queried test point; the nearest neighbor lookup dominates the computational cost of prediction, which limits the speedup precomputation provides.

We also consider embedding the 400 dimensional data in a lower dimensional feature space to ameliorate the prediction slowdown caused by the nearest neighbor search.
We use principal component analysis (PCA) to project the 400-dimensional image vectors into a 60-dimensional space (60 principal components capture 99 percent of the variance in the data).
We use an FCNN in place of a convolutional model for this experiment, as the embedding destroys the local structures within the data by orthogonal projection.

\begin{table}[h]
  \centering
\begin{tabular}{ |p{2.3cm}|p{1.6cm}|p{1.7cm}|p{1.3cm}|p{1.6cm}| p{1.3cm}|  }
  \hline
      & MuyGPs & FastMuyGPs & FCNN & SVGP & SGPR \\
  \hline
  Accuracy &  \textcolor{red}{$\bm{99.38}$} & 96.99 &  86.44   & 85.80   & 75.55 \\
  \hline
  Predict Time (s)  &  1.98e-1  & 7.16e-2 & \textcolor{red}{$\bm{1.89e}$$\bm{-2}$}  & 1.11e-1 & 6.37e-2 \\
  \hline
 \end{tabular}
 \newline
 \caption{Accuracy and prediction time on the embedded, $60$-dimensional  star-galaxy data. Highlighting represents the best performance in each statistic.}\label{table:star_galaxy_embedded}
\end{table}

Table~\ref{table:star_galaxy_embedded} shows that the fastest of the methods is the FCNN, while the most accurate is again MuyGPs. 
In this case, FastMuyGPs achieves better accuracy than that of the neural network (we expect that an FCNN should perform worse in image classification than a CNN), as well as SVGP and SGPR.
Comparing the non-embedded and embedded cases, we see that reducing the dimensionality of the data leads to improvement in accuracy for FastMuyGPs and MuyGPs, indicating that PCA transformation has regularizing properties from which the two approaches benefit.

\section{Conclusions, limitations, \& future work}\label{sec:future}
We have shown that FastMuyGPs achieves speedup over its predecessor MuyGPs in posterior mean prediction time for GPR.
Our approach features prediction speed close to that of deep neural networks and constant-time GP posterior mean prediction algorithms like SKI; it also enjoys many of the benefits of GPs, including their ability to learn latent functions in data-sparse applications.
FastMuyGPs is also accurate on the benchmark problems examined in this paper, maintaining the near-state-of-the-art performance MuyGPs has demonstrated in previous works.
A limitation of this approach is its inability to provide UQ of the predicted value in the form of a posterior variance. 
A future work could use some of the global kernel approximations mentioned in Section~\ref{sec:background} to 
obtain an approximation to the posterior variance of a given FastMuyGPs prediction in a memory-lean fashion. 
It would also be interesting to explore methods for subsampling the set of precomputed target vectors. 
Subsampling the precomputation would reduce the memory footprint of the algorithm and allow it to scale to even larger problems.
The current computational bottleneck in FastMuyGPs is the nearest neighbor lookup; optimizing that process would close the gap in prediction time between FastMuyGPs and methods like SKI and neural networks. 
Moreover, the success of FastMuyGPs on the borehole inverse problem, which does not require a posterior variance, suggests that FastMuyGPs will enable many related future applications.

\section{Societal impact}\label{sec:society}
GP regression methods have made tremendous impact in a variety of ML applications.
FastMuyGPs is well-suited to ML applications which benefit directly from fast predictions, such as RL and MCMC methods.
RL and MCMC algorithms are applied in fields as disparate as autonomous systems~\cite{ko2007gaussian} and healthcare~\cite{yu2021reinforcement}; these contributions have led to enhanced performance of self-driving cars and individual healthcare recommendations. 
Such advances can help address broad societal issues such as climate change through decreases in greenhouse gas emissions, public safety through reduction in traffic accidents, and public health through improvement in treatment and reduction in healthcare costs.
However, each of these advances is rife with risks for potential negative impacts.
Autonomous cars may have the potential to increase the safety of citizens; similar technology can be used in autonomous weapons systems such as battlefield drones.
Healthcare recommendation systems may improve outcomes for patients, but pose a privacy risk if data and recommendation models are not protected.
Such conflicts must be addressed and monitored to maximize the benefits of research in scalable GP algorithms while minimizing the negative impact of such research.
We therefore encourage research into the robustness of GP-based approaches within these applications, as well as a broader survey of all applications of GPs and the direct and indirect consequences of these applications.

\section{Acknowledgments}
This work was performed under the auspices of the U.S. Department of Energy by Lawrence Livermore National Laboratory under Contract DE-AC52-07NA27344 with IM release number LLNL-CONF-835226-DRAFT. 
Funding for this work was provided by LLNL Laboratory Directed Research and Development grant 22ERD028.
\bibliographystyle{plain}
\bibliography{paper}
\end{document}